\documentclass[letterpaper]{article} 
\usepackage{aaai2026}  
\usepackage{times}  
\usepackage{helvet}  
\usepackage{courier}  
\usepackage[hyphens]{url}  
\usepackage{graphicx} 
\usepackage{natbib}  
\usepackage{caption} 
\usepackage{algorithm}
\usepackage{algorithmic}

\usepackage{amssymb}
\usepackage{amsmath}
\usepackage{multirow}
\usepackage{amssymb}

\usepackage{subfig}

\usepackage{newfloat}
\usepackage{listings}
\DeclareCaptionStyle{ruled}{labelfont=normalfont,labelsep=colon,strut=off} 
\floatstyle{ruled}
\newfloat{listing}{tb}{lst}{}
\floatname{listing}{Listing}
\title{Fine-Grained Generalization via Structuralizing Concept and Feature Space into Commonality, Specificity and Confounding}

\author{
    Zhen Wang\textsuperscript{\rm 1},
    Jiaojiao Zhao\textsuperscript{\rm 1},  
    Qilong Wang\textsuperscript{\rm 2},
    Yongfeng Dong\textsuperscript{\rm 1}, 
    Wenlong Yu\textsuperscript{\rm 2,3}\thanks{Corresponding author.}
    }
\affiliations{
    \textsuperscript{\rm 1}School of Artificial Intelligence,
Hebei University of Technology, China \\
    \textsuperscript{\rm 2}Tianjin University, China \\
    \textsuperscript{\rm 3}Singapore Management University, Singapore \\
    \{wangzhen@, 202432803020@stu., dongyf@\}hebut.edu.cn, \{qlwang, ywl\_95\}@tju.edu.cn

}

\usepackage{bibentry}

\begin{document}

\maketitle

\begin{abstract}
Fine-Grained Domain Generalization (FGDG) presents greater challenges than conventional domain generalization due to the subtle inter-class differences and relatively pronounced intra-class variations inherent in fine-grained recognition tasks. Under domain shifts, the model becomes overly sensitive to fine-grained cues, leading to the suppression of critical features and a significant drop in performance. 
Cognitive studies suggest that humans classify objects by leveraging both common and specific attributes, enabling accurate differentiation between fine-grained categories. However, current deep learning models have yet to incorporate this mechanism effectively. Inspired by this mechanism, we propose Concept-Feature Structuralized Generalization (CFSG). This model explicitly disentangles both the concept and feature spaces into three structured components: common, specific, and confounding segments. 
To mitigate the adverse effects of varying degrees of distribution shift, we introduce an adaptive mechanism that dynamically adjusts the proportions of common, specific, and confounding components. 
In the final prediction, explicit weights are assigned to each pair of components. 
Extensive experiments on three single-source benchmark datasets demonstrate that CFSG achieves an average performance improvement of 9.87\% over baseline models and outperforms existing state-of-the-art methods by an average of 3.08\%. Additionally, explainability analysis validates that CFSG effectively integrates multi-granularity structured knowledge and confirms that feature structuralization facilitates the emergence of concept structuralization. 
\end{abstract}


\begin{links}
    \link{Code}{https://github.com/zhaozz-j/CFSG}
\end{links}

\section{Introduction}
Deep learning has recently led to significant advances in computer vision \cite{krizhevsky2017imagenet, 10889183} and natural language processing \cite{devlin2019bert}. 
However, these achievements rely primarily on the assumption that the training and testing data are drawn from the same distribution. In real-world scenarios, this assumption often fails due to distribution shifts across domains. 
For example, in image classification tasks, training data may consist of real-world bird images, while testing data could include artistic representations such as watercolor or oil paintings of birds. Due to these distributional differences across domains, traditional models often perform poorly on testing data. This issue is commonly referred to as the out-of-distribution (OOD) problem \cite{liu2021domain}. Domain Generalization (DG) aims to train deep learning models solely on source domain data to generalize effectively to OOD samples and has gained increasing attention in recent years \cite{wang2022generalizing}.

\begin{figure}[!t]
    \centering
    \includegraphics[width=1.0\linewidth]{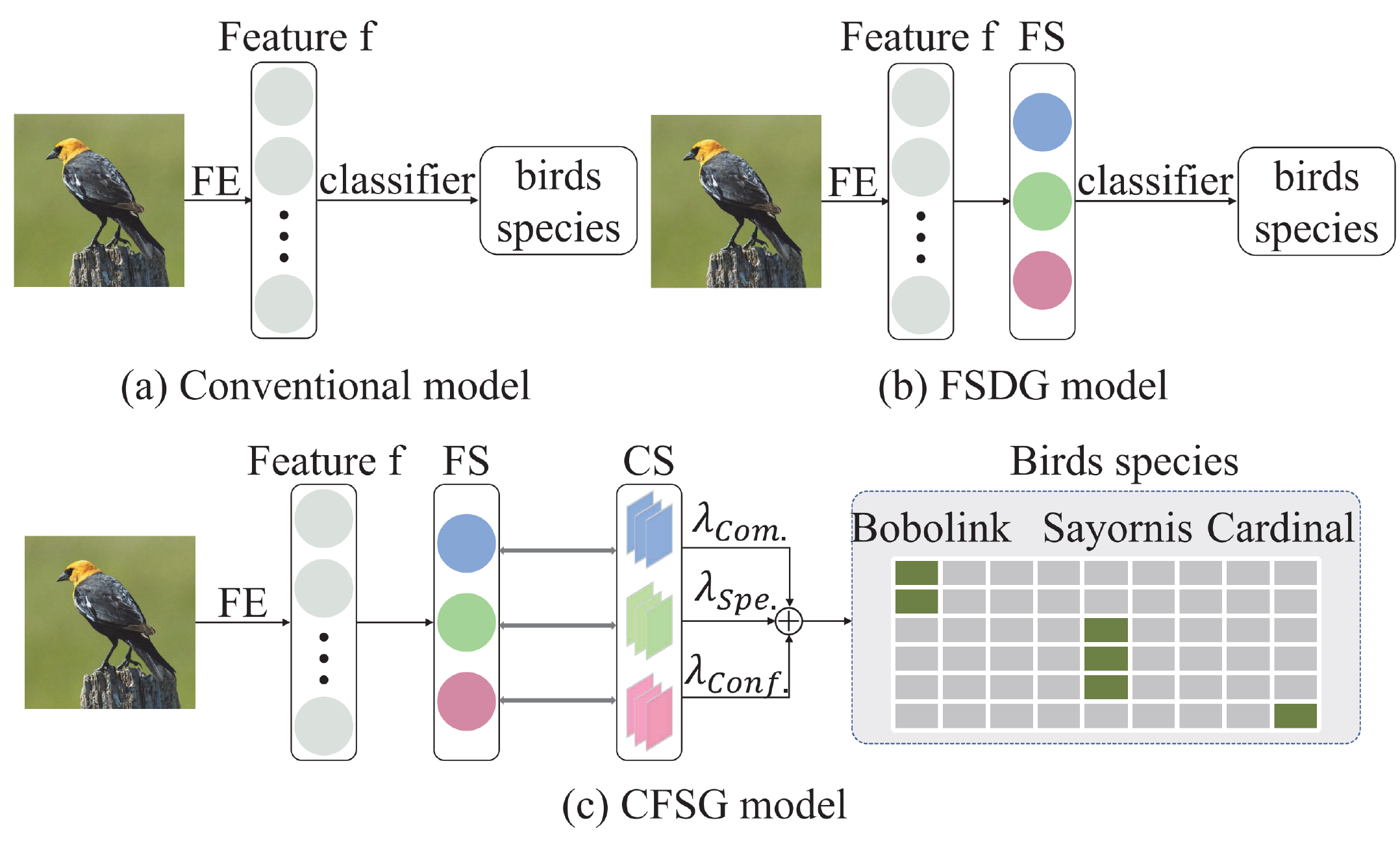}
    \caption{Decision mechanisms of different methods: conventional models predict directly from extracted features; FSDG structures the feature space, but classification still relies on features. In contrast, CFSG first achieves concept structuralization through structured features and then performs classification based on these structured concepts. FE stands for Feature Extractor, FS refers to Feature Structuralization, and CS represents Concept Structuralization.
    }
    \label{Fig1}
\end{figure}



Although some DG methods have made significant progress in addressing OOD challenges, their performance on fine-grained domain generalization (FGDG) tasks remains unsatisfactory \cite{wei2021fine, bi2025learning}. Fine-grained recognition focuses on subtle inter-class differences, such as variations among breeds within the same species. In these tasks, intra-class variations are often more pronounced than inter-class differences and may manifest differently across domains \cite{du2020fine}. Fine-grained tasks heavily rely on local discriminative regions; however, conventional techniques such as data augmentation damage these critical areas, leading to substantial performance drops. In addition, FGDG is particularly sensitive to cross-domain distribution shifts, posing greater challenges to traditional DG methods and limiting their generalization capabilities \cite{zhou2021domain}. Moreover, in real-world scenarios, conventional domain generalization methods, especially in single-source domain scenarios, tend to fail; the scarcity of multi-source fine-grained data makes single-source FGDG an even more demanding problem \cite{10485225}.

Humans significantly outperform machines on such tasks, primarily due to fundamental differences in decision-making processes. Cognitive studies indicate that humans do not rely solely on superficial similarity when distinguishing categories; instead, they leverage semantic concepts based on both commonality and specificity between categories, enabling robust generalization even when the target's appearance changes \cite{luppi2024information, yu2025coe,mervis1975family, mahner2025dimensions, du2025human, yu2025structural}. However, this decision-making strategy has yet to be effectively embedded in current deep learning models. Recently, Yu et al. \cite{yu2024fine} proposed an FSDG model that incorporates human decision-making mechanisms into the training process. By structuring the feature space and applying constraints, the model is guided to focus on both common and specific features, resulting in significantly improved generalization performance. However, this approach exclusively emphasizes structuralization in the feature space, while overlooking the equally important structural organization of the concept space. 

It is essential to integrate semantic concept learning into the decision-making process of deep learning models. Concept Bottleneck Models (CBMs) offer a representative approach by introducing a human-defined semantic concept bottleneck layer. The model first predicts these plain concepts under auxiliary supervision, and then makes the final decision based on them, leading to improvements in both performance and interpretability \cite{koh2020concept, he2025v2c}. Moreover, the Neural Collapse theory indicates that, under the influence of scaling factors, the classification layer weights in deep neural networks converge to the mean vectors of their respective classes, known as class concept prototypes \cite{papyan2020prevalence, song2024exploring}. Inspired by these works and cognitive science, this paper argues that deep networks should be structurally organized in both the concept and feature spaces to embed multi-granularity structured knowledge (shown in Fig. \ref{Fig1}).

We propose a novel domain generalization framework, Concept-Feature Structuralized Generalization (CFSG), that structurally disentangles the concept and feature spaces into three components: commonality, specificity, and confounding. Final classification is performed based on this structured learning and inference mechanism. 
Specifically, we first achieve concept space structuralization by disentangling the classification layer weights, and feature space structuralization by disentangling backbone-extracted features. Secondly, feature space structuralization is achieved by introducing multi-granularity structured constraints \cite{yu2024fine}. We further extends the Neural Collapse theory, proposing that structural constraints in the feature space also drive the concept space to exhibit similar structuralization, even in the absence of explicit semantic supervision. To achieve the third objective—classification based on structured features and concepts—this paper proposes adaptively mitigating the impact of varying degrees of distribution shifts on model performance by controlling the proportions of common, specific, and confounding components and explicitly assigning weights to each.
Following FSDG, we also disentangle the confounding component in the concept space to account for redundancy and randomness among category concepts. The main contributions of this paper are summarized as follows:
\begin{itemize}
    \item Drawing inspiration from human cognitive mechanisms, we propose the CFSG framework, which simultaneously and structurally decouples the concept and feature spaces into three components: commonality, specificity, and confounding.
    \item We argue that the impact of varying degrees of distribution shift on model performance can be adaptively mitigated by controlling the proportions of commonality, specificity, and confounding. To this end, we explicitly assign contribution weights to these three components during classification, thereby enhancing the model's generalization ability.
    \item Extensive experiments demonstrate that the proposed CFSG model achieves significant performance improvements on three single-source benchmark datasets, with an average gain of 9.87\% over baseline models and 3.08\% over state-of-the-art methods. Further explainability analysis reveals that CFSG effectively embeds multi-granularity structured knowledge and confirms that feature structuralization facilitates the emergence of concept structuralization.
\end{itemize}

\begin{figure*}[!t]
    \centering
    \includegraphics[width=1.0\linewidth]{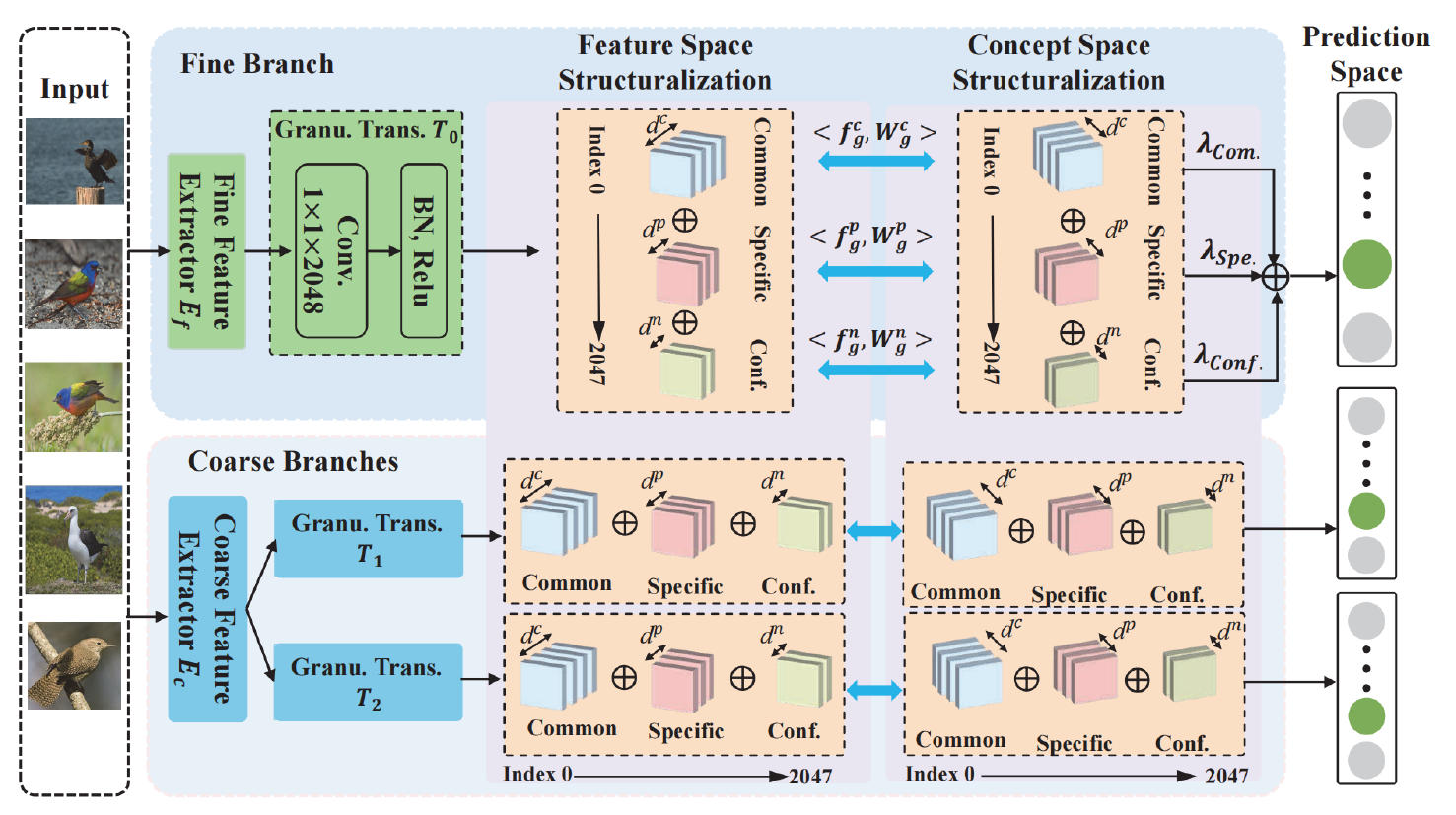}
    \caption{The model illustration presents an example based on a three-level granularity hierarchy. Granu. Trans. denotes the Granularity Transition Layer. Given an input image, CFSG performs structuralization and disentanglement in both the feature and concept spaces, and conducts classification by assigning different weights to the structured representations.}
    \label{Fig2}
\end{figure*}

\section{Related Work}
\textbf{Domain Generalization.} DG trains models solely on source domain data, aiming to improve generalization to OOD data. 
Deep learning has recently advanced under both independent and identically distributed (i.i.d.) and cross-domain settings \cite{10888795}, with DG receiving growing attention. Many methods have been proposed to improve DG, including domain alignment \cite{rosenfeld2021the}, meta-learning \cite{finn2017model}, and data augmentation \cite{yan2020improve}.


The core idea of domain alignment is to minimise discrepancies across multiple source domains to learn domain-invariant representations. 
DgCD \cite{kim2024discrepancy} measured inter-source domain differences and reduced reliance on domain-specific features via stochastic channel dropout. 
Data augmentation generated multiple virtual domains from a single source \cite{qiao2020learning}, improving cross-domain generalization.  
Zhou et al.  \cite{zhou2021domain} proposed the MixStyle method, which increased source domain diversity by mixing style information from training samples. Li et al. \cite{li2019feature} enhanced generalization by enabling the model to self-learn auxiliary loss functions.

However, these methods overlook the small inter-class and significant intra-class variations in fine-grained images. In addition, their reliance on multi-source data limits performance in single-source FGDG tasks.

\textbf{Fine-Grained Visual Categorization}. 
Fine-grained classification has garnered significant research attention in recent years \cite{xu2025context}.
Unlike general image recognition tasks, the goal of Fine-Grained Visual Classification (FGVC) is to distinguish subtle differences among subordinate categories within the same parent class, differences that are often imperceptible to the human eye \cite{wang2020dual, wei2021fine}. To address this issue, early studies primarily relied on additional annotations to assist classification through object localization \cite{berg2013poof}. However, such methods suffer from limited scalability. Over time, researchers shifted their focus to more scalable representation-learning frameworks \cite{chen2024fet}. Du et al. \cite{du2021progressive} proposed a progressive training strategy to integrate multi-granularity features, thereby improving fine-grained visual classification performance. 

These methods mainly focus on fine-grained recognition and do not adequately address domain shift. Achieving strong generalization in single-source FGDG remains a significant challenge, and existing approaches are still limited.
Currently, few methods have been proposed for single-source FGDG. Yu et al. \cite{yu2024fine} were the first to enhance cross-domain fine-grained semantic alignment by structuring the feature space. However, their approach overlooks the structuralization of the concept space, which is a crucial component. The HSSH framework incorporated a state-space model into the backbone network to enrich the style diversity of source domains, thereby improving generalization. However, it lacks a controllable mechanism during inference and applies only to the Vmamba architecture \cite{bi2025learning}. 

Unlike existing methods for addressing single-source FGDG, we draw inspiration from human cognitive mechanisms to simultaneously structure both the feature and concept spaces into common, specific, and confounding components. Furthermore, we believe that the impact of varying degrees of distribution shift on model performance can be adaptively mitigated by controlling the proportions of commonality, specificity, and confounding. 
Therefore, during classification, explicit weights are assigned separately to the three components.

\section{Methodology}
\subsection{Preliminaries}
Let $\mathcal{X}=\left\{(x)\right\}$ be the input space and $\mathcal{Y}=\left\{(y)\right\}$ the label space.
In the context of FGDG, data is constructed at multiple granularities. We focus exclusively on the single-source setting, where the source domain is defined as $\mathcal{D}_S = \left( x, y_f, y_g|_{g=1}^{G-1} \right)$. Specifically, the source domain consists of input data $x$ and corresponding labels $y = \left( y_f, y_g|_{g=1}^{G-1} \right)$, \(y_f\) represents the fine-grained labels, and \(y_g|_{g=1}^{G-1}\) represents the coarse-grained labels. In our method, data from a single source domain can generalize to multiple target domains, denoted as $\mathcal{D_T} = \left\{\mathcal{T}_k\right\}$. For different target domains, we assign distinct contribution weights to the structured common, specific, and confounding components during classification. Therefore, we aim to learn a model $f: \mathcal{X} \rightarrow \mathcal{Y}$ on the source domain data, to ensure that $f$ generalizes well to the target domain.

Fig.~\ref{Fig2} illustrates the overall framework of the proposed CFSG model. We jointly structuralize both the feature and concept spaces into three components: commonality, specificity, and confounding, and assign distinct classification weights to each structured pair of components.

\subsection{Concept-Feature Space Structuralization}
To embed the human decision-making mechanism based on commonality and specificity into deep learning models, we structurally disentangle the concept and feature spaces into three components: common, specific, and confounding. 
This disentanglement is performed along the channel dimension, with a fixed partition determined by channel indices and the proportion of each component.

Neural collapse theory indicates that, following proper scaling, the last-layer classifier weights converge to the class means, effectively representing the concept prototypes for each category \cite{papyan2020prevalence}. 
To go one step further, we propose that structurally disentangling features within deep learning models induces a corresponding disentanglement of the classifier weights, i.e., the concept prototypes of each class. In this work, we treat the classification layer weights as concept prototypes for each category and structurally disentangle the concept space into three components: common, specific, and confounding.

Specifically, in CFSG, data is organized in a multi-granularity format, where the number of categories varies across different granularities $g$. Consequently, the shape of the classification layer weight matrix $W_g$ differs across granularities $g$. We disentangle the weight matrix $W_g$ along the channel dimension into three components: common, specific, and confounding, with the partitioning fixed by channel indices and ratios. Formally defined as follows:
\begin{equation}
    {Disentangle (W_g) = \{W_g^c,W_g^p,W_g^n\}}.
\end{equation}
Here,
${W}_{{g}}^{c} = \{w_{g}^{c,i}\mid i = 1, 2, \ldots, d^c \}$
represents the \( d^{c}\) common parts,
${W}_{{g}}^{p} = \{w_{g}^{p,i}\mid i = 1, 2, \ldots, d^p \}$
reflects the specific parts, and 
${W}_{{g}}^{n} = \{w_{g}^{n,i}\mid i = 1, 2, \ldots, d^n \}$
denotes the confounding parts. Through the above approach, the concept space is structured into common, specific, and confounding components.

After structurally disentangling the concept space, we further disentangle the feature space. Specifically, given a batch of source images \(x\in R^{B\times3\times W\times H}\), we adopt both coarse-grained and fine-grained feature extractors to capture fine-grained semantic information better and ensure a fair comparison with FSDG. For comparison, we also conduct experiments using a single shared feature extractor. After passing through two feature extractors, \( E_ c\) and \( E_ f\), the coarse-grained features \( F_ c\) and fine-grained features \( F_ f\) are obtained, respectively. Subsequently, these features pass through a Granularity Transition Layer (GTL), which selects the relevant features for each granularity. This yields a feature set \( F_ g\),  composed of individual features 
 \(\{f_g^1,f_g^2,\ldots,f_g^d\}\), where $d$ denotes the number of channels. We then disentangle these features along the channel dimension into three components, following the same approach as in the concept space. The components are formally defined as:
\begin{equation}
    {Disentangle (F_g) = \{F_g^c,F_g^p,F_g^n\}}, 
\end{equation}
where
${F}_{{g}}^{c} = \{f_{g}^{c,i}\mid i = 1, 2, \ldots, d^c \}$
represents the \( d^{c}\) common features,
${F}_{{g}}^{p} = \{f_{g}^{p,i}\mid i = 1, 2, \ldots, d^p \}$
reflects the specific parts, and 
${F}_{{g}}^{n} = \{f_{g}^{n,i}\mid i = 1, 2, \ldots, d^n \}$
denotes the confounding parts, satisfying \({d}=d^c+d^p+d^n\). 

With the above design, we achieve structural disentanglement of both the concept space and the feature space simultaneously. The key to achieving structuralization lies in disentanglement followed by alignment. Inspired by neural collapse theory \cite{papyan2020prevalence}, we propose that concept structuralization can be realized through constraints imposed by feature structuralization, such that the alignment constraints for concepts follow the same principles as those for features. Specifically, we compute the similarity between the common, specific, and confounding features and their corresponding concepts, respectively.

\subsection{Concept Bottleneck-Based Classification}
After achieving structuralization in both the concept and feature spaces, classification is performed based on the structured feature and concept representations. We propose that adaptively controlling the proportions of common, specific, and confounding components can mitigate the impact of varying degrees of distribution shifts on model performance by assigning explicit weights to each component.

Specifically, our classifier differs from conventional ones. Traditional deep learning models typically employ a parameterized softmax classifier, which is generally formulated as: $H=(W^T\times f_{\theta}(x))+b$. Here, \(f_{\theta}\) denotes the feature extracted from the input image \(x\), \(W\) is the weight matrix of the fully connected layer, and \(b\) is the bias term. Both \(W\) and \(b\) are learnable parameters optimized during training. 

In CFSG, our classifier is defined as follows:
\begin{align}
    H =\ & \lambda_c \langle f^c_g, W^c_{k,g} \rangle + \lambda_p \langle f^p_g, W^p_{k,g} \rangle \nonumber \\
          & + \lambda_n \langle f^n_g, W^n_{k,g} \rangle + b, 
\end{align}
where $\langle \cdot \, , \cdot \rangle$  denotes the inner product. \(f_g^c\), \(f_g^p\), and \(f_g^n\) represent the features of the sample after decoupling along the channel dimension through the GTL. \(W_{k,g}^c\), \(W_{k,g}^p\), and \(_{k,g}^n\) denote the classifier weights corresponding to the commonality, specificity, and confounding components, respectively, for category $k$ at granularity $g$. \(\lambda_c\), \(\lambda_p\), and \(\lambda_n\) represent the explicit weighting coefficients for the three parts, indicating their relative contributions in the weighted aggregation of distances between the sample features and the respective sub-weight components. During inference, the weights of the three components are normalized to sum to one and are manually adjusted to achieve optimal generalization.


\subsection{Feature Alignment Constraint}
During training, we adopt the standard $\mathcal{L}_{FSDG}$ loss from the FSDG \cite{yu2024fine} method as our objective function, which is defined as follows:
\begin{equation} \label{zongloss}
    \mathcal{L}_{FSDG} = \mathcal{L}_{c} + \mathcal{L}_{lf} + \mathcal{L}_{FS}.
\end{equation}
The detailed explanation is as follows:
\begin{itemize}
    \item $\mathcal{L}_{c}$: The coarse-grained cross-entropy loss, defined as $\mathcal{L}_{c} = \sum_{g=1}^{G-1} L_{CE}(\widehat{y}_g, y_g)$.
    \item $\mathcal{L}_{lf}$: The prediction alignment loss, where fine-grained ground-truth labels are progressively fused with the coarse-grained branch's prediction distribution. This fusion leverages coarse-grained information to enhance the performance of the fine-grained branch, thereby achieving effective alignment in the prediction space.
    \item $\mathcal{L}_{FS}$: The constraint loss from the feature structuralization module, encompassing feature disentanglement $\mathcal{L}_{dec}$ and the feature alignment loss $\mathcal{L}_{fa}$. Formally, $\mathcal{L}_{FS} = \mathcal{L}_{dec} + \mathcal{L}_{fa}$. The feature alignment loss $\mathcal{L}_{fa}$ enforces constraints on both common and specific features, expressed as $\lambda_{p}\mathcal{S}_{p} - \lambda_{cs}\mathcal{S}_{cs} - \lambda_{cd}\mathcal{S}_{cd}$.
\end{itemize}
We further optimize the loss function and training procedure of FSDG and adopt a more advanced inference strategy during model prediction.

For the feature extractor, we instantiate the backbone using networks from the ResNet (RN) series \cite{he2016deep}, Vision Transformer (ViT) series \cite{dosovitskiy2021an}, or ASMLP series \cite{lian2022asmlp}. Additionally, to enable fair comparison with the HSSH method, we adopt the Vmamba \cite{liu2024vmamba} series as the backbone on the CUB-Paintings and Birds-31 datasets. 
Since the official HSSH code has not been released, we are unable to conduct comparisons on the CompCars dataset. We use fine-grained classification accuracy as the evaluation metric. All experiments are performed on an A30 GPU. At deployment, only the fine-grained branch is used, while the coarse-grained branch serves solely as an auxiliary module during training.

\section{Experiments}

We evaluate CFSG on three fine-grained datasets and compare it with various state-of-the-art DG and FGDG methods, demonstrating its robustness across multiple mainstream backbone networks.

\subsection{Datasets}

\begin{table}[!t]
	\centering
	{
	\begin{tabular}{c|ccc|c}
		\hline
		Method & C $\rightarrow$ P &  P $\rightarrow$ C & Avg & Params\\ 
        \cline{1-5}
        \textit{ResNet-Based:} & \\
        PAN(DA) & 67.40 & 50.92 & 59.16 & 103M \\
        ERM & 54.94 & 35.67 & 45.31 & 24M \\
        ARM & 47.98 & 31.53 & 39.76 & 24M \\
        DANN & 54.06 & 37.09 & 45.57 & 24M \\
        MLDG & 55.40 & 34.15 & 44.78 & 23M \\
        GroupDRO & 54.94 & 35.67 & 45.31 & 23M \\
        CORAL & 54.70 & 35.29 & 45.00 & 23M \\
        SagNet & 56.33 & 36.71 & 46.52 & 24M \\
        MixStyle & 52.97 & 28.44 & 40.71 & 23M \\
        Mixup & 54.58 & 34.66 & 44.62 & 23M \\
        RIDG & 36.41 & 24.11 & 30.26 & 24M \\
        SAGM & 57.83 & 37.16 & 47.50 & 23M \\
        MIRO & 56.29 & 41.28 & 48.79 & 47M \\
        \hline
        S-FSDG & 63.42 & 44.87& 54.14& 26M\\
        \textbf{S-CFSG} (Ours) & \underline{64.06} & \underline{54.06} & \underline{59.06} & 28M\\
        FSDG & 61.84 & 49.46 & 55.65 & 26M \\
        \textbf{CFSG} (Ours) & \textbf{67.50} & \textbf{65.62} & \textbf{66.56} & 28M \\
        \hline
       \textit{VMamba-Based:} & \\
        HSSH  & \underline{67.48} & \underline{64.57} & \underline{66.03} & 35M \\
        \textbf{CFSG} (Ours) & \textbf{70.94} & \textbf{73.75}& \textbf{72.35}& 35M\\
        \hline
	\end{tabular}
        }
        \caption{Classification accuracy (\%) of the proposed CFSG method and existing methods on the Cub-Paintings dataset. The best results are highlighted in bold, and the second-best is underlined. }
	\label{tab:cub}
\end{table}

\begin{table}[!h]
	\centering
	{
	\begin{tabular}{c|ccc|c}
		\hline
		Method &  W $\rightarrow$ S &  S $\rightarrow$ W & Avg & Params\\ 
        \cline{1-5}
        \textit{ResNet-Based:} & \\
        PAN (DA) & 47.05 & 15.57 & 31.31 & 103M \\
        ERM      &  44.15 &  7.54 &  25.85 & 24M \\
        ARM      &  20.25 &  4.74 &  12.50 & 24M \\
        DANN     &  35.10 &  6.80 &  20.95 & 24M \\
        MLDG     &  44.94 &  7.56 &  26.25 & 23M \\
        GroupDRO &  43.60 &  7.75 &  25.68 & 23M \\
        CORAL    &  43.05 &  7.97 &  25.51 & 23M \\
        SagNet   &  45.33 &  8.89 &  27.11 & 24M \\
        MixStyle &  38.37 &  6.28 &  22.33 & 23M \\
        Mixup    &  43.07 &  7.56 &  25.32 & 23M \\
        RIDG     &  36.57 &  8.11 &  22.34 & 24M \\
        SAGM     &  49.55 &  8.58 &  29.07 & 23M \\
        MIRO     &  46.01 &  7.88 &  26.95 & 47M \\
        \hline
        S-FSDG & 53.44 & 10.83 & 32.14& 26M\\
        \textbf{S-CFSG} (Ours) & \underline{73.12} & \underline{17.19} & \underline{45.16}& 28M\\
        FSDG & 51.78 & 11.30 & 31.54 & 26M \\
        \textbf{CFSG} (Ours) & \textbf{74.69} & \textbf{20.62} & \textbf{47.66} & 28M\\
        \hline
	\end{tabular}
        }
        \caption{Classification accuracy (\%) of the proposed CFSG method and existing methods on the ComCars dataset. The best results are highlighted in bold, and the second-best is underlined.}
	\label{tab:cars}
\end{table}

\begin{table*}[!t]
	\centering
	\resizebox{0.80\linewidth}{!}{
	\begin{tabular}{c|ccccccc|c}
		\hline
		Method & C $\rightarrow$ I &  C $\rightarrow$ N &  I $\rightarrow$ C &  I $\rightarrow$ N &  N $\rightarrow$ C &  N $\rightarrow$ I & Avg & Params\\ 
        \cline{1-9}
        \textit{ResNet-Based}: & \\
        PAN (DA) & 69.79 & 84.19 & 90.46 & 88.10 & 92.51 & 75.03 & 83.34 & 103M \\
        ERM &  54.64 & 72.93 & 85.01 & 74.97 & 86.10 & 62.51 & 72.69 & 24M \\
        ARM & 50.51 & 71.25 & 77.38 & 74.20 & 84.74 & 59.82 & 69.65 & 24M \\
        DANN & 52.75 & 71.82 & 80.79 & 73.59 & 85.55 & 61.53 & 71.01 & 24M \\
        MLDG & 53.55 & 72.19 & 80.74 & 74.83 & 85.61 & 61.95 & 71.48 & 23M \\
        GroupDRO & 52.61 & 70.78 & 81.87 & 74.40 & 86.26 & 61.32 & 71.21 & 23M \\
        CORAL & 54.64 & 72.93 & 81.01 & 74.97 & 86.10 & 62.51 & 72.03 & 23M \\
        SagNet & 53.66 & 71.75 & 81.39 & 74.13 & 85.66 & 62.06 & 71.44 & 24M \\
        MixStyle & 49.95 & 69.04 & 74.46 & 6834 & 83.60 & 57.12 & 67.09 & 23M \\
        Mixup & 52.36 & 71.65 & 82.36 & 75.17 & 85.61 & 62.34 & 71.58 & 23M \\
        RIDG & 47.15 & 66.71 & 82.47 & 73.63 & 85.77 & 60.98 & 69.45 & 24M \\
        SAGM & 54.04 & 73.63 & 82.96 & 77.01 & 87.88 & 63.49 & 73.17 & 23M \\
        MIRO & 54.39 & 74.87 & 82.36 & 75.34 & 86.42 & 62.48 & 72.64 & 47M \\
        \hline
        S-FSDG & 63.66 & \underline{82.43} & 89.56& 85.80& 92.03& 72.91& 81.06 & 26M\\
        \textbf{S-CFSG} (Ours) & \underline{70.94} & 76.88 & 90.63 & 84.69 & \underline{94.06} & \underline{75.63} & 82.14 & 28M\\
        FSDG & 66.32 & \textbf{83.71} & \underline{90.96} & \underline{87.36} & 91.95 & 74.20 & \underline{82.37} & 26M \\
        \textbf{CFSG} (Ours) & \textbf{72.81} & 80.94 & \textbf{94.06} & \textbf{88.13} & \textbf{95.31} & \textbf{78.44} & \textbf{84.95}& 28M\\
        \hline
        \textit{VMamba-Based:} & \\
        HSSH  & \underline{80.43} & \textbf{90.39} & \underline{95.25} & \textbf{94.33} & \underline{96.17} & \textbf{87.57} & \textbf{90.69} & 35M \\
        \textbf{CFSG} (Ours) & \textbf{81.25} & \underline{85.31}& \textbf{98.12} & \underline{93.44} & \textbf{99.06} & \underline{85.94}& \underline{90.52}& 35M\\
        \hline
	\end{tabular}
        }
        \caption{Classification accuracy (\%) of the proposed CFSG method and existing methods on the Birds-31 dataset. The best results are highlighted in bold, and the second-best is underlined.}
	\label{tab:bd}
\end{table*}

\textbf{CUB-Paintings} consists of two distinct domains: CUB-200-2011 (C) \cite{wah2011caltech} and CUB-200-Paintings (P) \cite{wang2020progressive}. Both datasets follow a four-level fine-grained classification hierarchy, encompassing 14 orders, 38 families, 122 genera, and 200 species. CUB-200-2011 contains 11,788 real bird images, whereas CUB-200-Paintings includes 3,047 cross-media artistic renditions.

\textbf{CompCars} \cite{yang2015large} includes car images from two distinct sources: the web (\textbf{W}) and surveillance (\textbf{S}). It is organized into a hierarchical classification system with 68 coarse categories and 281 fine-grained categories.

\textbf{Birds-31} consists of three domains: CUB-200-2011 (\textbf{C}), NABirds (\textbf{N}) \cite{van2015building}, and iNaturalist2017 (\textbf{I}) \cite{van2018inaturalist}. In \cite{wang2020progressive}, the categories from these three datasets were united, and 31 fine-grained categories were selected, with corresponding image counts of 1,848, 2,988, and 2,857. A four-level granularity structure was subsequently constructed, including 4 orders, 16 families, 25 genera, and 31 species.

\subsection{Experimental Results}
We evaluate the effectiveness of the proposed method by comparing it with a variety of DG and FGDG approaches, including ERM \cite{erm}, ARM \cite{zhang2021adaptive}, DANN \cite{ganin2016domain}, MLDG \cite{li2018learning}, GroupDRO \cite{sagawa2019distributionally}, CORAL \cite{sun2016deep}, SagNet \cite{nam2021reducing}, MixStyle \cite{zhou2021domain}, Mixup \cite{yan2020improve}, RIDG \cite{chen2023domain}, SAGM \cite{wang2023sharpness},
MIRO \cite{cha2022domain}, FSDG \cite{yu2024fine}, and HSSH \cite{bi2025learning}. All methods are implemented based on their official code, except for HSSH. Since the official HSSH code is not publicly available, we compare our method against the experimental results reported in its original paper. 
We evaluate the proposed CFSG method under both dual-backbone and single-backbone (prefixed with "S-") training settings. 

\textbf{On the CUB-Paintings dataset,} as shown in Table \ref{tab:cub}, our method achieves the best performance among all  DG competitors. The dual-backbone and single-backbone models outperform the baseline FSDG model by 10.91\% and 4.92\%, respectively. 
When replacing the backbone with the Vmamba series, CFSG outperforms HSSH by 6.32\%. 
These results significantly confirm the effectiveness of our proposed CFSG model, which performs classification based on structured feature and concept representations while adaptively weighting the contributions of the three components.

\textbf{On the ComCars dataset,} as shown in Table \ref{tab:cars}, our method demonstrates significant improvement. The dual-backbone and single-backbone models achieve improvements of 16.12\% and 13.02\% over the baseline FSDG model. These results demonstrate that our proposed method effectively addresses distribution shifts between source and target domains and highlights the critical role of concept structuralization in enhancing generalization.

\textbf{On the Birds-31 dataset}, as shown in Table \ref{tab:bd}, our method achieves the highest average accuracy, exhibiting a clear performance improvement. Under the dual-backbone configuration, it surpasses FSDG by 2.58\%, while in the single-backbone setting, it outperforms the baseline by 1.08\%. When replacing the backbone network with the Vmamba series, the CFSG model achieves performance comparable to the HSSH method, demonstrating the robustness of our approach across different backbone architectures. In contrast, HSSH shows superior performance only on the Vmamba architecture.

\textbf{Experiments on various backbone architectures with different depths.} To verify the robustness of CFSG across different architectures, we conduct experiments on the feature extractors within the model pipeline using various backbone networks and deep architectures, including RN-50, RN-101, ViT-Tiny, ViT-Small, ASMLP-Tiny, and ASMLP-Small. The detailed experimental results are presented in Table \ref{tab:backbone}. Experimental results show that our method consistently improves performance across various architectures, with average performance gains ranging from 2.05\% to 11.19\%. These results demonstrate the robustness of our method across different backbone networks.


\begin{table}[!h]
    \centering
    \resizebox{\linewidth}{!}{
    \begin{tabular}{c|c|ccc|c}
        \hline
        Backbone & Method & C $\rightarrow$ P & P $\rightarrow$ C & Avg & Params\\ 
        \hline
        \multirow{2}{*}{RN-101} 
            & FSDG & 64.64 & 49.86 & 57.25 & 45M \\
            & CFSG & 65.62 & 71.25 & 68.44 & 47M \\
        \hline
        \multirow{2}{*}{ViT-T} 
            & FSDG & 60.71 & 48.82 & 54.76 & 11M \\
            & CFSG & 62.50 & 52.50 & 57.50 & 13M \\
        \hline
        \multirow{2}{*}{ViT-S} 
            & FSDG & 70.26 & 66.25 & 68.26 & 31M \\
            & CFSG & 70.31 & 70.31 & 70.31 & 33M \\
        \hline
        \multirow{2}{*}{MLP-T} 
            & FSDG & 60.90 & 50.51 & 55.70 & 31M \\
            & CFSG & 63.13 & 59.06 & 61.10 & 33M \\
        \hline
        \multirow{2}{*}{MLP-S} 
            & FSDG & 63.67 & 53.84 & 58.75 & 53M \\
            & CFSG & 67.50 & 66.87 & 67.19 & 55M \\
        \hline
    \end{tabular}
    }
    \caption{Classification accuracy (\%) of the proposed CFSG method on the Cub-Paintings dataset using various backbones with different depths. RN and MLP denote ResNet and ASMLP backbones, respectively. T and S stand for Tiny and Small, respectively.}
    \label{tab:backbone}
\end{table}

\begin{table}[!t]
\centering
\begin{tabular}{ccc|ccc}
\hline
Backbone & CS & FS & $\mathrm{C} \rightarrow \mathrm{P}$ & $\mathrm{P} \rightarrow \mathrm{C}$ & Avg \\
\hline
$\checkmark$ &  &  & 56.76 & 46.53 & 51.65 \\
$\checkmark$ &  & $\checkmark$ & 61.84 & 49.46 & 55.65 \\
$\checkmark$ & $\checkmark$ &  & 57.50 & 60.00 & 58.75 \\
$\checkmark$ & $\checkmark$ & $\checkmark$ & \textbf{67.50} & \textbf{65.62} & \textbf{66.56} \\
\hline
\end{tabular}
\caption{Classification accuracy (\%) of different modules in the CFSG model, where CS denotes Concept Structuralization and FS denotes Feature Structuralization.}
\label{tab:CF-structuralization}
\end{table}

\subsection{Analysis}
\textbf{Parameter Impact Analysis.} CFSG adaptively mitigates the impact of distribution shifts of varying degrees on model performance by controlling the proportions of common, specific, and confounding components. To validate the effectiveness of our method, we conduct extensive experiments to explore the existence of an optimal balance strategy. The results indicate the existence of such an optimum from domain P to domain C, with the best weights being 0.75, 0.2, and 0.05, respectively. We visualize different weight configurations and perform experiments treating the weights as learnable parameters. More implementation details are provided in the Appendix.

\textbf{Concept Structuralization and Feature Structuralization Modules.} We evaluate the effects of concept structuralization only, feature structuralization only, and their combination on top of the backbone network. When the feature structuralization module is not used, the model employs the standard cross-entropy loss. As shown in 
Table~\ref{tab:CF-structuralization} shows that feature structuralization alone improves performance by 4.0\%, concept structuralization by 7.1\%, and their combination achieves the highest improvement of 14.91\%. These results further validate the effectiveness of jointly structuralizing both the concept and feature spaces.

\begin{table}[!t]
\centering
\begin{tabular}{cccc}
\hline
Method & C $\rightarrow$ P &  P $\rightarrow$ C & Avg\\
\hline
FSDG & 61.84 & 49.46 & 55.65\\
CFSG(sub-centroids) & 62.81 & 56.25 & 59.53\\ 
CFSG & \textbf{67.50} &  \textbf{65.62} & \textbf{66.56}\\ 
\hline
\end{tabular}
\caption{Classification accuracy (\%) of the sub-centroid-based method on the CUB-Paintings dataset.}
\label{tab:sub-centroid}
\end{table}

\textbf{Compared to methods that perform classification by constructing sub-centroids.} According to the Neural Collapse theory, classifier weights converge to class feature means up to a scaling factor \cite{papyan2020prevalence}. Inspired by this, we assume that the structuralized classifier weights also converge to sub-centroids. Based on this assumption, we design a sub-centroid-based classification method and compare it with CFSG. During training, sub-centroids are constructed, and during inference, classification is performed by computing the weighted distances between samples and the sub-centroids. Detailed construction procedures are provided in the Appendix. Table \ref{tab:sub-centroid} reports the experimental results. The sub-centroids method based on sample means achieves a 3.88\% improvement over the FSDG baseline, demonstrating its effectiveness. However, its performance remains inferior to that of the CFSG method. Classification based on concept structuralization achieves a substantial performance boost.

\textbf{Explainability analysis.} We posit that, under multi-granularity structured knowledge, categories that are closer in distance exhibit higher similarity in the concept space. We measured both the ground-truth category similarity and the concept space similarity, then computed the Spearman rank correlation between them. As shown in Table \ref{tab:Ground Truth}, the CFSG model achieves a high correlation of 0.97, significantly outperforming the unstructured FGDG baseline with 0.69. This result confirms that CFSG effectively embeds multi-granularity knowledge into both concept and feature spaces, and that feature structuralization facilitates the emergence of concept structuralization. More implementation details are provided in the Appendix.

\begin{table}[!t]
\centering
\begin{tabular}{c   |cccc}
\hline
FGDG & All & Com. & Spe. & Conf.\\
\hline
Ground Truth & 0.69 & 0.70 & 0.67 & 0.67 \\
\hline
CFSG & All & Com. & Spe. & Conf.\\
\hline
Ground Truth & 0.97 & 0.97 & 0.97 & 0.91 \\
\hline
\end{tabular}
\caption{Spearman's rank correlation between region-level concepts and ground-truth labels for each model.}
\label{tab:Ground Truth}
\end{table}

\section{Conclusion}
The core contribution of this work lies in interpreting classifier weights as an unsupervised concept space and proposing that structured patterns in the feature space can induce structural organization in the concept space. We jointly structuralize both the feature and concept spaces into three components: common, specific, and confounding. By controlling the proportions of common, specific, and confounding parts, the model adaptively mitigates the impact of varying degrees of distribution shifts on performance. During classification, the model computes the similarity between structured features and structured concept representations, assigning explicit weights to each component. Extensive experiments demonstrate that the proposed CFSG method significantly improves performance in FGDG scenarios.

\section{Acknowledgments}
This work was supported in part by the National Natural Science Foundation of China (NSFC) under Grants 62306103, 62276186, and 62376194, and in part by the Higher Education Science and Technology Research Project of Hebei Province under Grant QN2023262.

\bibliography{aaai2026}

\newpage
\section{Appendix}

\subsection{Neural Collapse}
Deep learning models often exhibit an inductive bias known as Neural Collapse during the late stages of training \cite{papyan2020prevalence}. This phenomenon encompasses four highly correlated behaviors. The first states that, for a given class, the deep features (i.e., the activations from the final layer) of all training samples collapse to their class mean. This behavior can be formally expressed as:
\begin{equation}
\forall i \in \mathcal{S}_c, \quad {h}_{i,c} = {\mu}_c + {\epsilon}_{i,c}, \quad \text{where } {\epsilon}_{i,c} \rightarrow {0}, 
\end{equation}
$\mathcal{S}_c$ denotes the set of all samples belonging to class $c$, and ${\mu}_c$ represents the mean feature vector of class $c$.

The second phenomenon of Neural Collapse is that the distances between each class center and the global mean of all samples are equal, and these class centers converge to the vertices of an equiangular tight frame simplex. It can be formally defined as:
\begin{equation}
    \forall c, c' \in \{1, \dots, C\}, \quad \| {\mu}_c - {\mu}_G \| \rightarrow \| {\mu}_{c'} - {\mu}_G \|, 
\end{equation}
${\mu}_c$ and ${\mu}_{c'}$  denote the class centers of classes $c$ and ${c'}$,respectively, while ${\mu}_G$ represents the global center of all samples.

The third phenomenon of Neural Collapse states that, under a scaling factor, the classifier weights of a deep learning model converge to the class means. Formally, this is represented as:
\begin{equation}
    \frac{\mathbf{w}_c}{\|\mathbf{w}_c\|} \rightarrow \frac{\mathbf{\mu}_c}{\|\mathbf{\mu}_c\|}. 
\end{equation}
The fourth phenomenon of Neural Collapse indicates that the final classification decision of a deep learning model reduces to selecting the class whose prototype is closest to the activation vector. The formulation can be written as:
\begin{equation}
    \mathop{\arg\min}\limits_{c \in \{1,\dots,C\}} \|{h} - \mathbf{\mu}_c\| \rightarrow \mathop{\arg\max}\limits_{c \in \{1,\dots,C\}} {w}_c^\top {h} + b_c. 
\end{equation}
Inspired by the phenomenon of neural collapse, this work treats the classification layer weights as an unsupervised concept space. Furthermore, we propose that feature structuralization induces concept structuralization, thereby simultaneously decomposing both the concept and feature spaces into common, specific, and confounding components. This explicit integration of human decision-making mechanisms into deep learning models enhances fine-grained generalization.

\subsection{Experimental details}
During training, we adopt the loss function $\mathcal{L}_{FSDG}$ proposed in FSDG \cite{yu2024fine} as our objective, which is defined as follows:
\begin{equation} \label{zongloss1}
    \mathcal{L}_{FSDG} = \mathcal{L}_{c} + \mathcal{L}_{lf} + \mathcal{L}_{FS}, 
\end{equation}
here, $\mathcal{L}_{c}$ denotes the coarse-grained cross-entropy loss. All coarse branches are also trained to minimize the standard classification objective, which is defined as:
\begin{equation}
\mathcal{L}_{c} = \sum_{g=1}^{G-1} L_{CE}(\widehat{y}_g, y_g).
\end{equation}

$\mathcal{L}_{lf}$ denotes the prediction alignment loss. The prediction alignment loss is designed to progressively fuse fine-grained ground-truth labels with the prediction distribution of the coarse-grained branch during training, thereby bridging the prediction spaces of both granularities and optimizing the fine-grained branch. The formulation of $\mathcal{L}_{lf}$ is given by:
\begin{equation} 
\label{llh}
\begin{aligned}
    \mathcal{L}_{lf} (\left.\widehat{y}_g  \right| &  _{g=1}^{G-1}, \widehat{y}_f, \left.  y_f\right) \\ &
    = D_{\mathrm{KL}}\left(\varepsilon y_f +
    (1-\varepsilon) \sum_{g=1}^{G-1} \frac{\widehat{y}_g}{G-1} \| \widehat{y}_f\right),
\end{aligned}
\end{equation}
$\hat{y}_f$ denotes the fine-grained predicted distribution, $D_{\mathrm{KL}}$ represents the Kullback–Leibler divergence, $\hat{y}_g$ is the output of the coarse-grained branch, and $\varepsilon$ controls the influence of the coarse-grained prediction on the fine-grained branch.

$\mathcal{L}_{FS}$ represents the feature space constraints, encompassing both the feature disentanglement loss and the alignment loss. The loss $\mathcal{L}_{FS}$ is formally expressed as:
\begin{equation} \label{lfs}
\begin{aligned}
    \mathcal{L}_{FS} & = \mathcal{L}_{dec} + \mathcal{L}_{fa}\\ &
    = \mathcal{L}_{dec} - \mathcal{L}_{cs} - \mathcal{L}_{cd} + \mathcal{L}_{p}\\
    &
    = \mathcal{L}_{dec} - \lambda_{cs}\mathcal{S}_{cs} - \lambda_{cd}\mathcal{S}_{cd} + \lambda_{p}\mathcal{S}_{p},
\end{aligned}
\end{equation}
$\mathcal{L}_{dec}$ denotes the disentanglement loss, which aims to ensure the independence and orthogonality among the three components. It is formally defined as:
\begin{equation}
\label{l_dec}
    \mathcal{L}_{dec} = \frac{1}{B}\frac{1}{G}\sum_{b=0}^{B-1}\sum_{g=0}^{G-1}SUM(
    {\left\langle P_{b,g}^{all}, P_{b,g}^{all}\right\rangle 
    \over \Vert P_{b,g}^{all} \Vert_2 \Vert P_{b,g}^{all} \Vert_2} - {I}).
\end{equation}
$P_{b,g}^{all}$ represents the stacked tensor of the three feature components for each sample. The disentanglement along the channel dimension follows a fixed ratio of 5:3:2 for common, specific, and confounding components, respectively. 

$\mathcal{L}_{fa}$ denotes the feature alignment loss. $\mathcal{S}_{cs}$ is the first constraint applied to the common features, encouraging consistency of the same sample's common representations across adjacent granularities. We employ cosine similarity to measure this consistency, which is defined as follows:
\begin{equation} 
\label{cs}
    S_{cs} =  \frac{1}{B}\frac{1}{G-1}\sum_{b=0}^{B-1}\sum_{g=0}^{G-2}SUM({
    \left\langle{f}_{b,g}^{c},{f}_{b,g+1}^{c}\right\rangle 
    \over \Vert {f}_{b,g}^{c} \Vert_2 \Vert {f}_{b,g+1}^{c} \Vert_2}
    ),
\end{equation}
${f}_{b,g}^{c}$ denotes the common feature component of the $b$-th image at granularity level $g$. $\mathcal{S}_{cd}$ serves as another constraint on the common features. $\mathcal{S}_{cd}$ is an additional constraint on the common features, encouraging consistency among the common representations of subcategories that share the same parent class. It is defined as:
\begin{equation} 
\label{cd}
    S_{cd} =
    \frac{1}{G-1}\frac{1}{Q_g}\frac{1}{K}\sum_{g=0}^{G-2}\sum_{q=0}^{Q_g-1}SUM({\left\langle P_{q,g}^c, P_{q,g}^c\right\rangle 
    \over \Vert P_{q,g}^c \Vert_2 \Vert P_{q,g}^c \Vert_2} - {I}
    ), 
\end{equation}
$P_{q,g}^c$ represents the common feature prototype of subcategories under the parent category $q$ at granularity level $g$. Here, $Q_g$ denotes the number of parent categories at this granularity level, and $K$ denotes the total number of subcategories. Both of these constraints on common features aim to maximize similarity; therefore, a negative sign is applied when incorporating them into the overall loss function. $\lambda_{cs}$ and $\lambda_{cd}$ are the coefficients of the two corresponding constraints in the overall loss function. $\mathcal{S}_{p}$ is a constraint imposed on the specific features, defined as:
\begin{equation} 
\label{p}
    S_{p} =  \frac{1}{G-1}\frac{1}{K}\sum_{g=0}^{G-1}SUM({{\left\langle P_{g}^p, P_{g}^p\right\rangle 
    \over \Vert P_{g}^p \Vert_2 \Vert P_{g}^p \Vert_2} - {I}}
    ), 
\end{equation}
$P_{g}^p$ denotes the prototype representation of category-specific features at granularity level $g$, where $K$ is the number of subcategories at this granularity. This constraint encourages the specific features of different categories at the same granularity level to be as distinct as possible. $\lambda_{p}$ is the coefficient of the specificity feature constraint in the overall loss function.

We further optimize the loss function and training procedure of FSDG and adopt a more advanced inference strategy during model prediction. Features extracted by the backbone network are passed through a Granularity Transition Layer (GTL) to obtain features corresponding to each granularity level. This layer consists of a convolutional layer, a batch normalization layer, and a ReLU activation function. Additionally, to unify the output dimensions of different backbone networks, we insert a convolutional layer with 2048 output channels between the backbone and the granularity transformation layer. During CFSG training, the model can adopt either a dual-backbone or single-backbone architecture. At deployment, only the fine-grained branch is used, while the coarse-grained branch serves as an auxiliary module during training. Except for the backbone network, all other layers are trained from scratch, with a fixed batch size of 32. We use fine-grained classification accuracy as the evaluation metric. During each training run, the model is evaluated on the target domain ten times. 

The proposed CSFG method is orthogonal to existing DG and FGDG approaches, allowing seamless integration with current generalization-enhancing methods (e.g., HSSH \cite{bi2025learning}) to further improve performance.

\subsection{Experimental Analysis}
\begin{figure}[!t]
    \centering
    \includegraphics[width=1.0\linewidth]{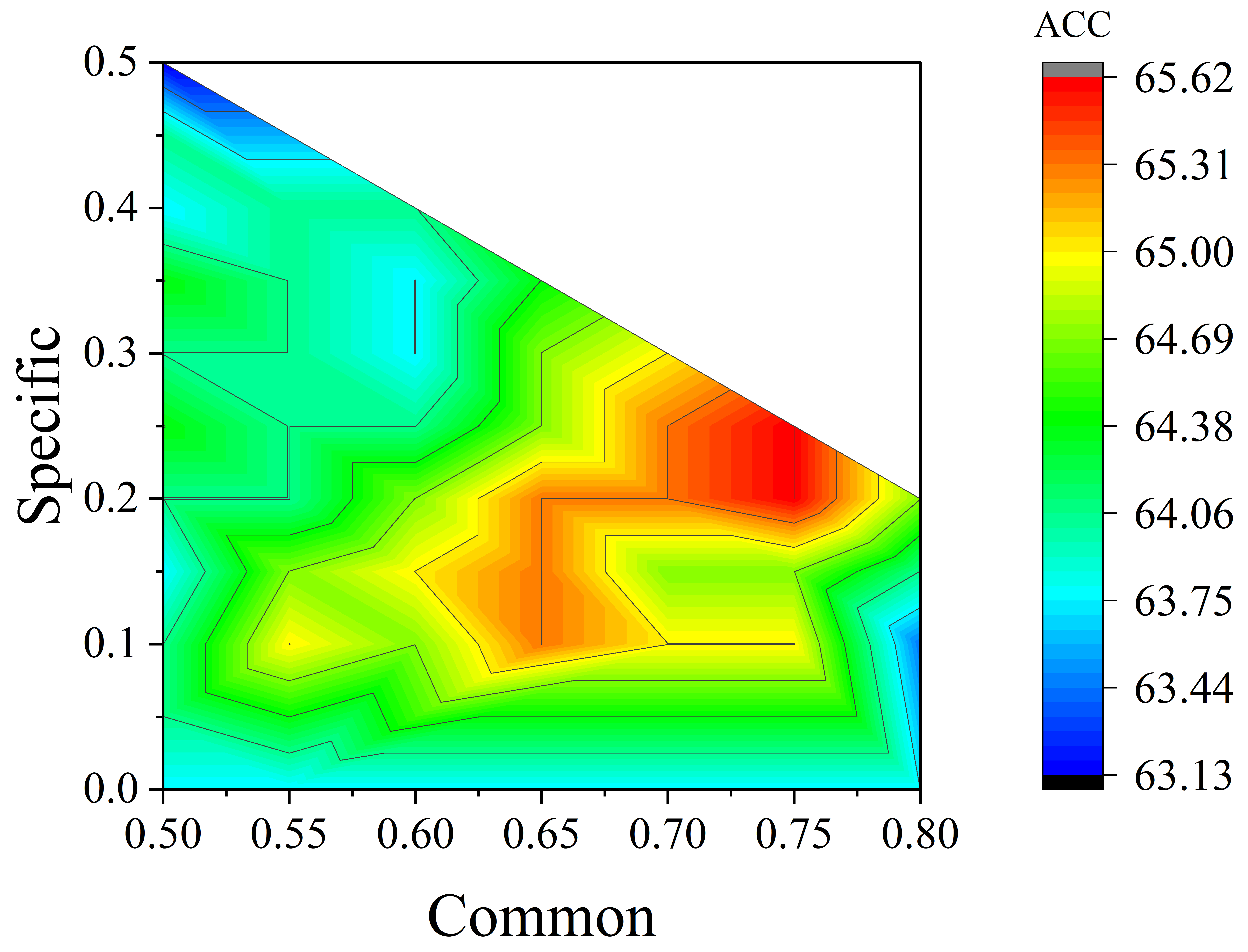}
    \caption{Classification accuracy (\%) of CFSG under varying commonality, specificity, and confounding weights when trained on P and tested on C.}
    \label{Fig4}
\end{figure}

\textbf{Parameter Impact Analysis.} To mitigate the impact of varying degrees of distribution shift, CFSG adaptively adjusts the proportions of common, specific, and confounding components. To validate this mechanism, we conduct extensive experiments to investigate whether an optimal weighting strategy exists for counteracting specific types of distribution shift. In the dual-backbone setting, we conduct a series of experiments on the CUB-Paintings dataset using different weighting strategies during training and inference. Specifically, during training, all component weights are set to 1. Inference, various weight combinations are tested, with the sum of the three component weights constrained to 1, to evaluate their impact on generalization performance from the P domain to the C domain. Results show that the optimal weights are 0.75, 0.2, and 0.05 for the common, specific, and confounding components, respectively. We visualize the performance under different configurations of commonality, specificity, and confounding weights, as shown in Fig.~\ref{Fig4}.

We further explore a learnable variant where the model automatically optimizes the three parameters. The experimental results, shown in Table \ref{tab:study}, indicate that using the same parameter settings during both training and inference—i.e., allowing the model to learn the weights for the three components autonomously—yields significantly worse performance than our explicitly weighted approach. Analysis of the learned weights for commonality, specificity, and confounding components reveals that the model excessively emphasizes commonality while assigning near-zero weights to specificity and confounding components, ultimately leading to performance degradation.

\begin{table}[!t]
\centering
\begin{tabular}{cccc}
\hline
Method & C $\rightarrow$ P &  P $\rightarrow$ C & Avg\\
\hline
CFSG(Learnable Weights) & 61.06 &  48.94 & 55.00\\ 
CFSG(Fixed Weights) & 67.50 & 65.62 & 66.56\\
\hline
\end{tabular}
\caption{Classification accuracy (\%) of CFSG under fixed and learnable weight settings on the CUB-Paintings dataset.}
\label{tab:study}
\end{table}

\textbf{Evaluation in both in-domain and out-of-domain scenarios.} We evaluate the accuracy of the proposed method and baseline models under both in-domain and out-of-domain settings, with detailed results provided in Table \ref{tab:domain}. All models are trained on the CUB-200-2011 dataset. In-domain performance is evaluated on its test set, while out-of-domain performance is assessed on the CUB-200-Painting dataset. Experimental results show that the CFSG model incurs a performance drop of approximately 5.43\% under the in-domain setting, but achieves a 0.64\% improvement in the out-of-domain setting. These results validate that CFSG enhances generalization by structurally disentangling both the concept and feature spaces, effectively reducing the model's reliance on domain-dependent characteristics.

\begin{table}[!t]
\begin{tabular}{ccc}
\hline
Method & In-Dom. Acc. & Out-of-Dom. Acc. (C $\rightarrow$ P)\\
\hline
S-FSDG & 86.37 & 63.42\\ 
S-CFSG & 80.94 &  64.06\\ 
\hline
\end{tabular}
\caption{Classification accuracy (\%) of in-domain and out-of-domain scenarios.}
\label{tab:domain}
\end{table}

\begin{figure*}[!th]
\centering

\begin{tabular}{cccc}
    \subfloat[All of FGDG]{\includegraphics[width=0.24\textwidth]{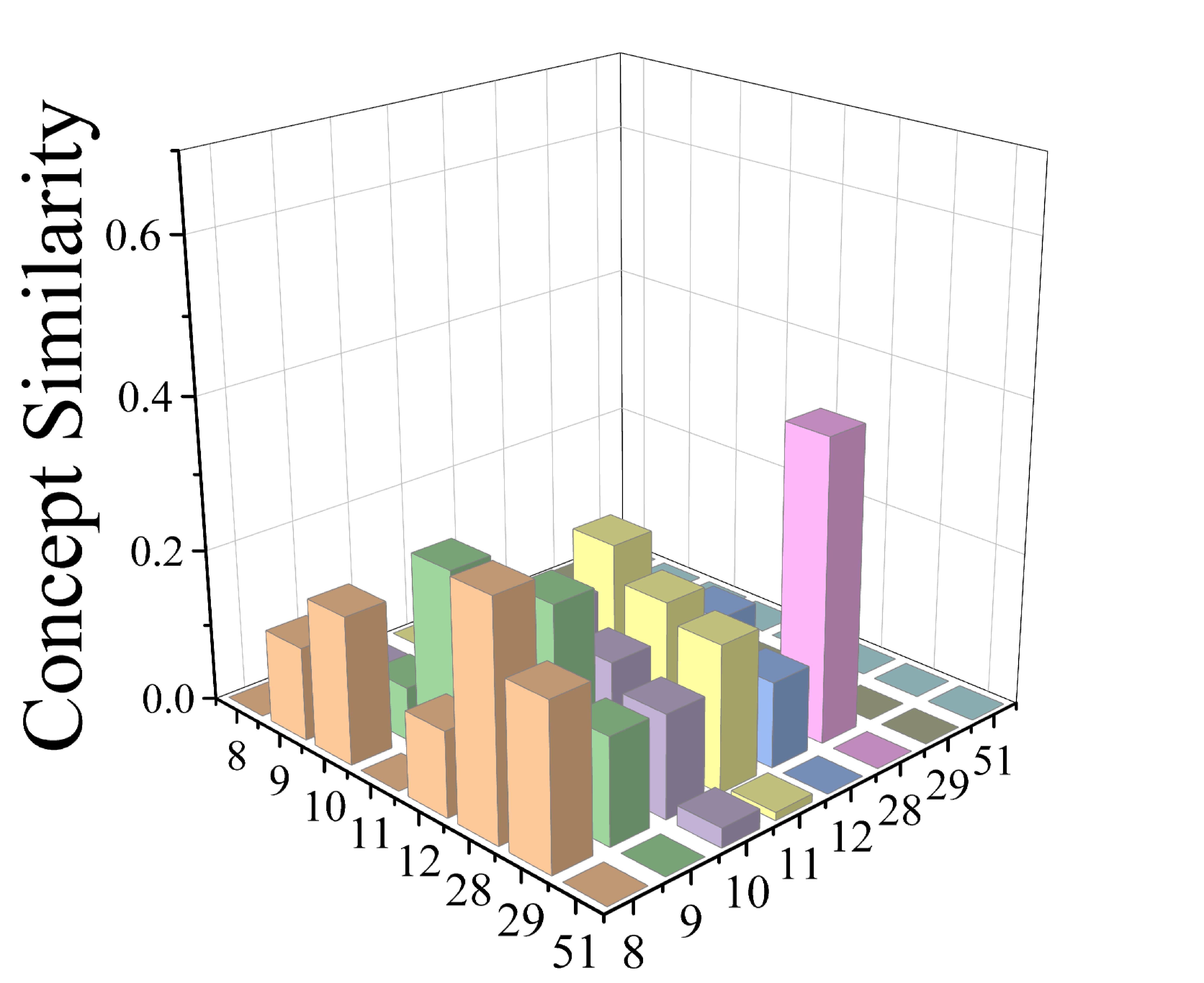}\label{allfgdg}} &
    \subfloat[Common of FGDG]{\includegraphics[width=0.24\textwidth]{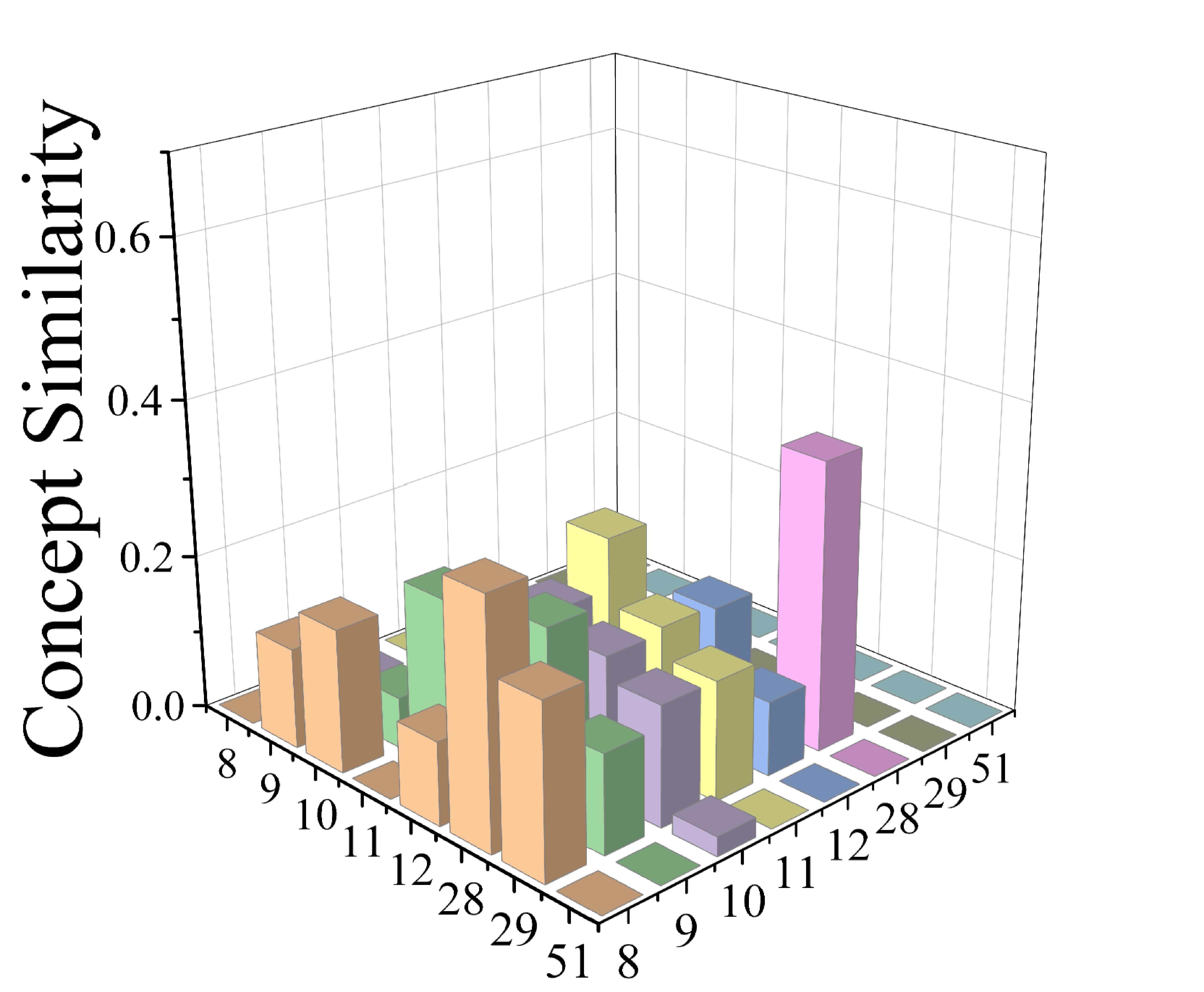}\label{comfgdg}} &
    \subfloat[Specific of FGDG]{\includegraphics[width=0.24\textwidth]{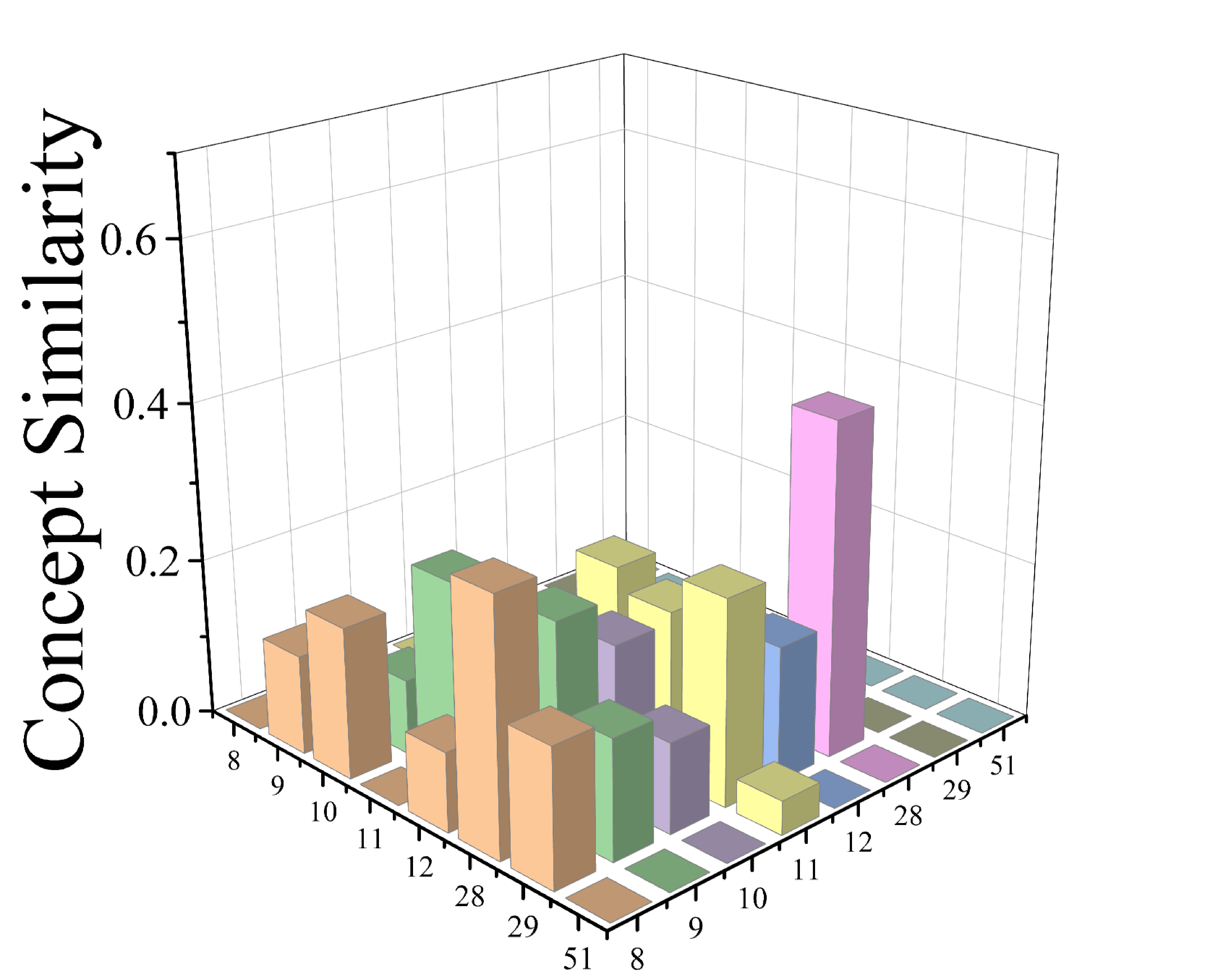}\label{spefgdg}} &
    \subfloat[Confounding of FGDG]{\includegraphics[width=0.24\textwidth]{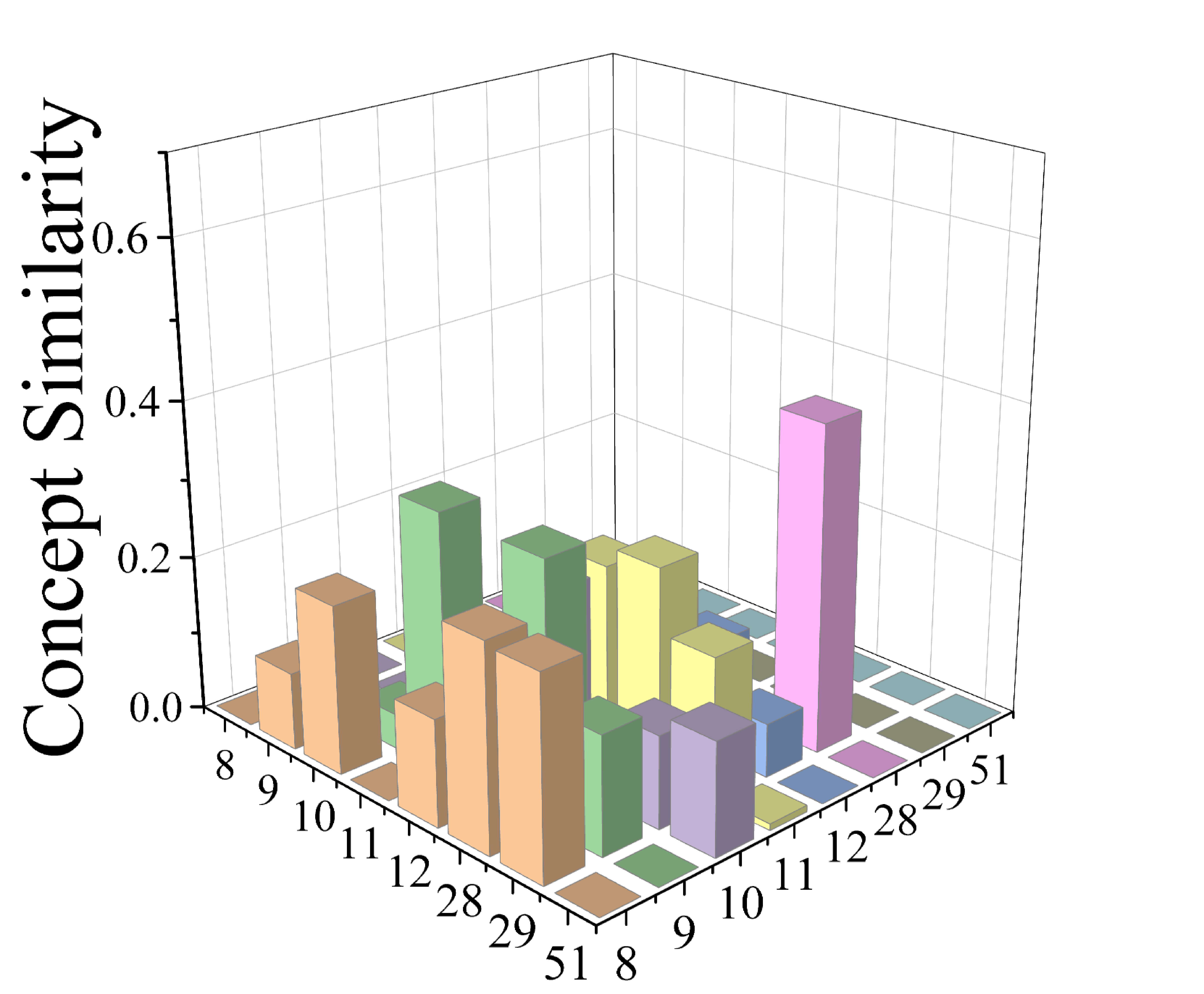}\label{noifgdg}} \\

    \subfloat[All of CFSG]{\includegraphics[width=0.24\textwidth]{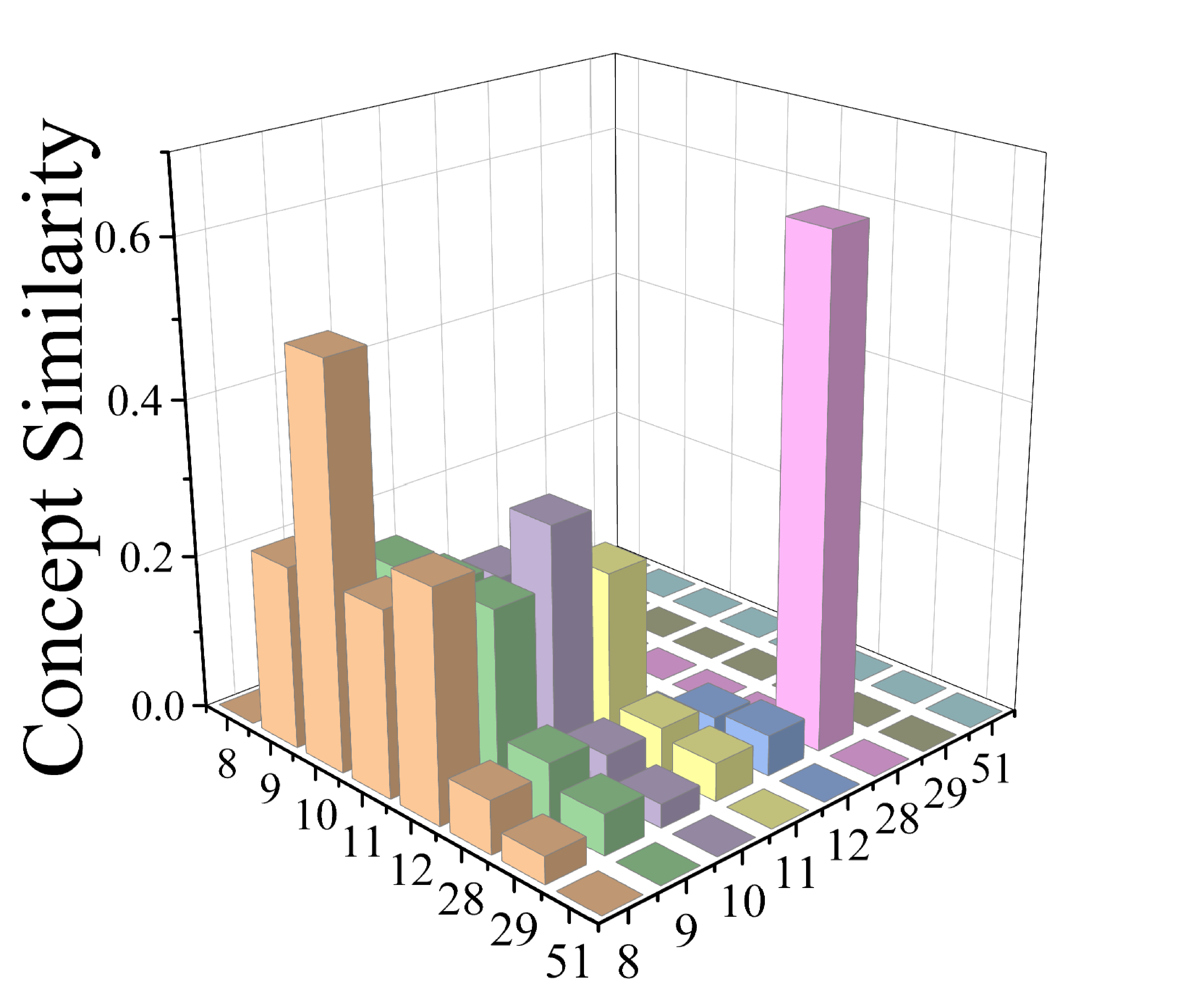}\label{allCFSG}} &
    \subfloat[Common of CFSG]{\includegraphics[width=0.24\textwidth]{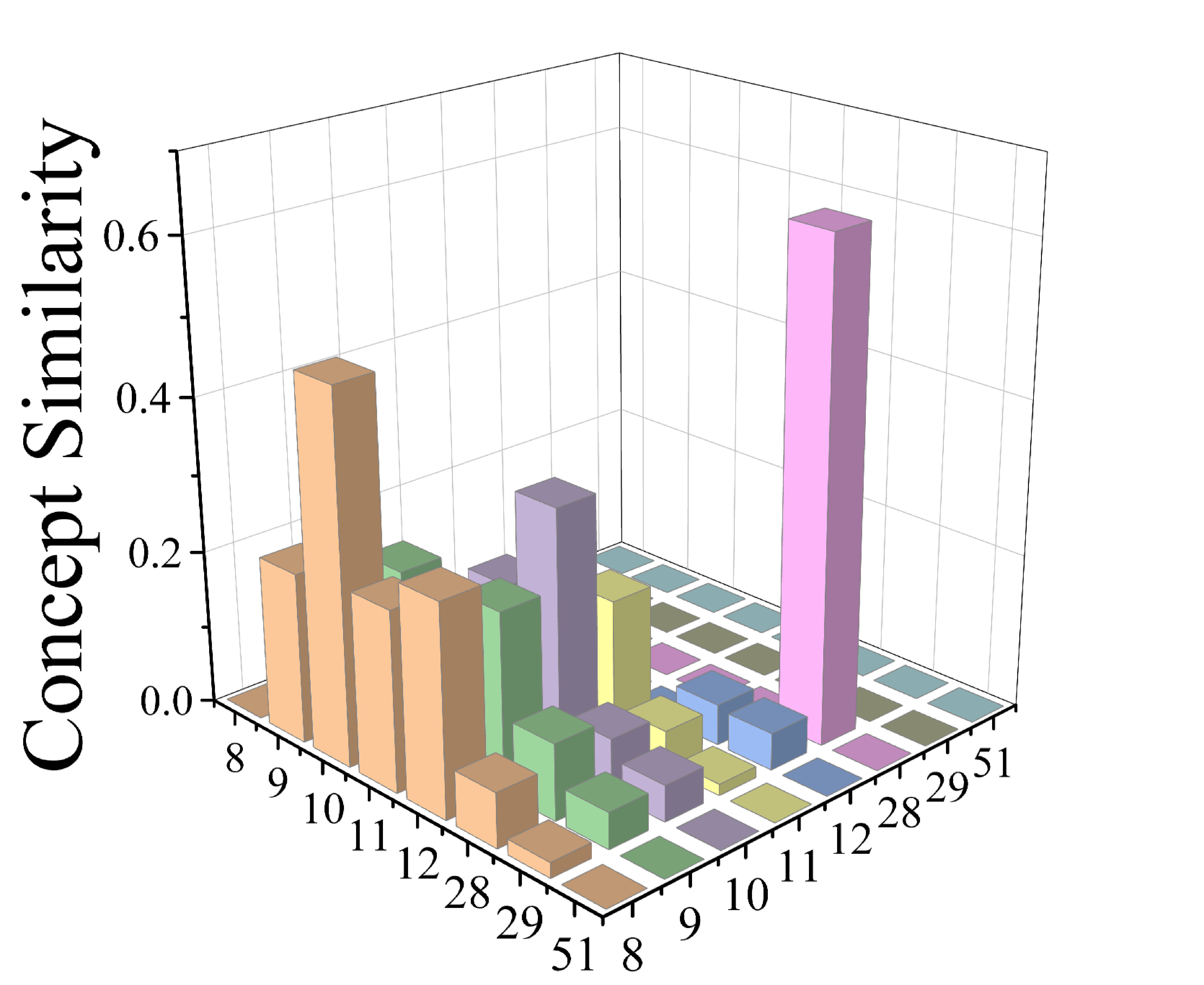}\label{comCFSG}} &
    \subfloat[Specific of CFSG]{\includegraphics[width=0.24\textwidth]{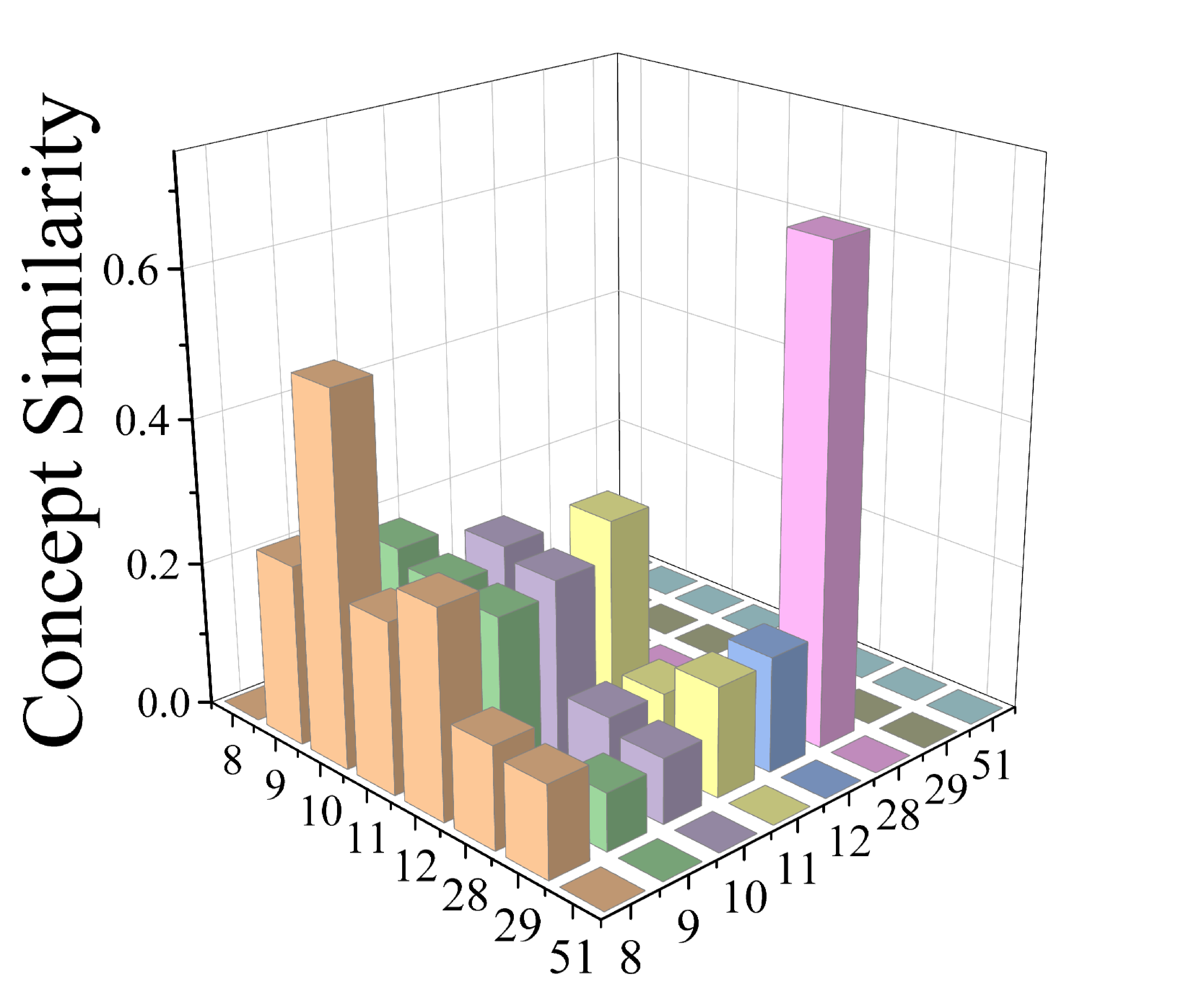}\label{speCFSG}} &
    \subfloat[Confounding of CFSG]{\includegraphics[width=0.24\textwidth]{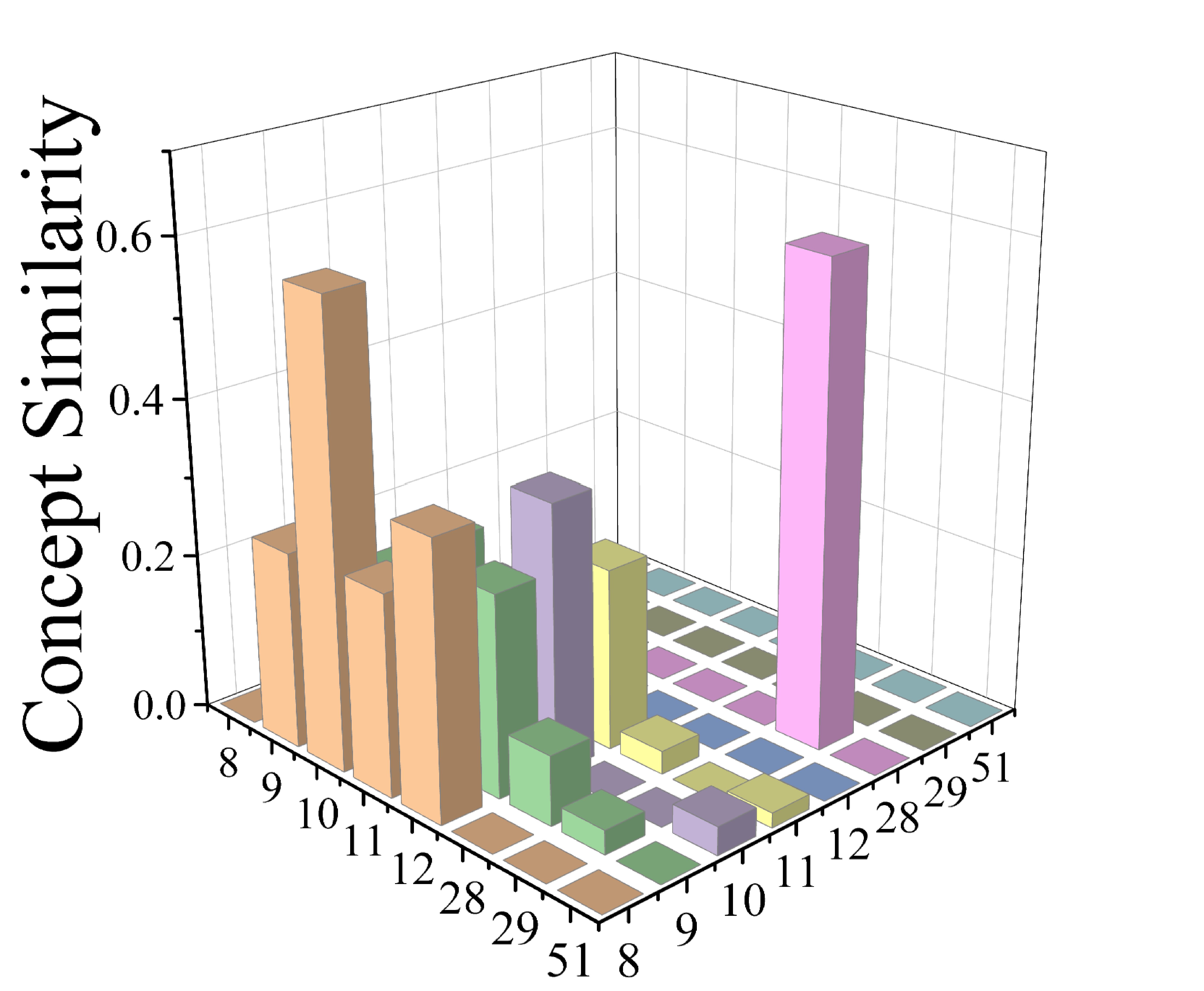}\label{noiCFSG}}
\end{tabular}

\caption{Cosine similarity results of FGDG and CFSG across common, specific, and confounding concepts.}
\label{fig:alloverlap}
\end{figure*}

\subsection{Constructing sub-centroids}
\label{sec:constructing sub-centroids}
For each class-specific sub-centroid, we first compute the class prototype from the current mini-batch:
\begin{equation}
\label{Pg}
\begin{array}{l}
P_g^{c} = MEAN(MEAN(F_g^c, {dim}=2), {dim}=0) \\
P_g^{n} = MEAN(MEAN(F_g^n, {dim}=2), {dim}=0) \\
P_g^{p} = MEAN(MEAN(F_g^p, {dim}=2), {dim}=0), 
\end{array}
\end{equation}
$F_g^c$, $F_g^n$, and $F_g^p$ denote the common, specific, and confounding feature components, respectively, obtained by structurally disentangling the feature representations along the channel dimension through the GTL. The prototype features for the classes in the current batch are obtained by averaging the corresponding structured features over the spatial (height × width) and sample dimensions. 
Subsequently, we update the sub-centroids of each class using a momentum-based approach:
\begin{equation}
\label{Pg}
\begin{array}{l}
F_{k,g}^{c}=\mu\times F_{k,g}^c+(1-\mu)\times P_{k,g}^c\\
F_{k,g}^p=\mu\times F_{k,g}^p+(1-\mu)\times P_{k,g}^p\\
F_{k,g}^n=\mu\times F_{k,g}^n+(1-\mu)\times P_{k,g}^n, 
\end{array}
\end{equation}
$F_{k,g}^{c}$, $F_{k,g} {p} $, and $F_{k,g} {n} $ represent the common, specific, and confounding sub-centroids of class $k$ at granularity $g$, respectively. Here, $\mu$ denotes the momentum parameter used for updating. After obtaining the sub-centroids for each class during training, the inference stage performs classification based on the distance between the sample and the sub-centroids. It is formally defined as:
\begin{equation}
\begin{array}{rl}
\hat{y} = \arg\min_k \Big( 
& \lambda_c \| f_g^c - F_{k,g}^c \|_2 + \lambda_p \| f_g^p - F_{k,g}^p \|_2 \\
& + \lambda_n \| f_g^n - F_{k,g}^n \|_2 \Big), 
\end{array}
\end{equation}
$f_g^c$, $f_g^p$, and $f_g^n$ denote the common, specific, and confounding features of the inference sample, respectively, after structural disentanglement via the GTL. We determine the class of each sample by computing the Euclidean distance to the corresponding sub-centroids. To adaptively mitigate the impact of varying degrees of distribution shifts on model performance, we assign different weights to the common, specific, and confounding components within the overall distance calculation by adjusting their respective proportions.

\subsection{Explainability analysis}
\label{sec:Explainability analysis}
In this work, we propose that structural patterns in the feature space can induce structural organization in the concept space. We treat the classifier weights as an unsupervised concept space and jointly structuralize both the feature and concept spaces. By classifying based on structured features and conceptual prototypes, our method achieves significant improvements in FGDG. In this section, we aim to evaluate the alignment between the concept space and the ground-truth class labels.

\begin{figure}[!t]
    \centering
    \includegraphics[width=1.0\linewidth]{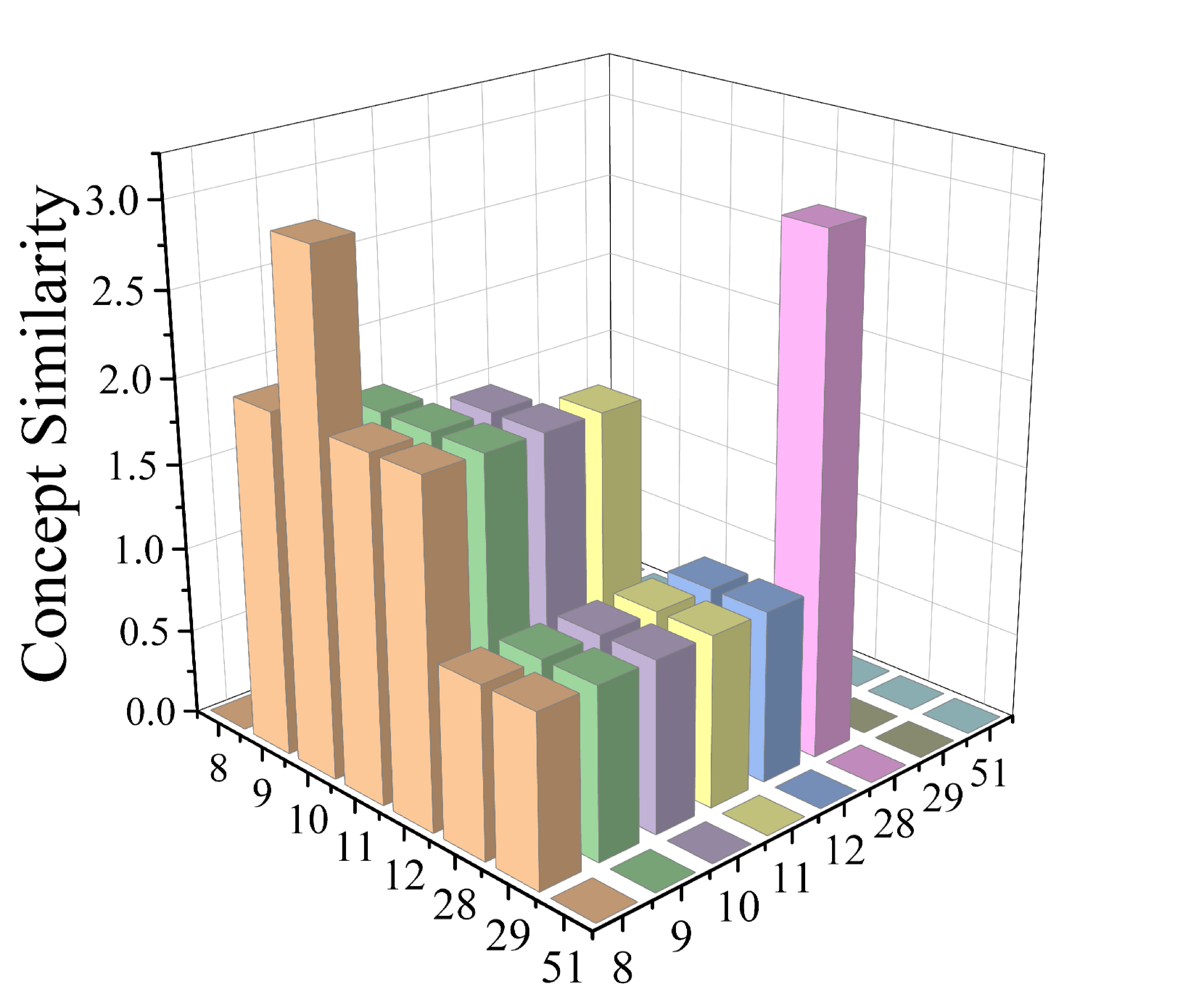}
    \caption{Confusion matrix of the ground truth for concept similarity.}
    \label{truth}
\end{figure}

We argue that category pairs with closer label distances tend to exhibit higher similarity in the concept space, whereas those with greater label distances show lower similarity. To validate this hypothesis, we selected a subset of categories from the CUB-Paintings dataset, along with their corresponding labels at four granularity levels, summarized in Table \ref{tab:label}. Following the multi-granularity structure-based distance metric defined in FSDG \cite{yu2024fine}, we measured the similarity between fine-grained categories using the following approach:
\begin{equation} \label{sclass}
    {Similarity}_{class}^{i, j}=d_i-\left\|\boldsymbol{c}_i-\boldsymbol{c}_j\right\|_0,
\end{equation}
$\boldsymbol{c}_i$ denote the label vector of a category, and $d_i$ represent the dimension of $\boldsymbol{c}_i$. We calculate the similarity between each pair of fine-grained categories and construct a confusion matrix, as shown in the fig .\ref{truth}.

To further quantify inter-class similarity in the concept space, we computed the similarity between all category concept prototypes. The similarity is calculated according to the following expression:
\begin{equation}
\text{cosine\_similarity}({W}_k, {W}_{k+1}) = \frac{{W}_k \cdot {W}_{k+1}}{\|{W}_k\| \|{W}_{k+1}\|}, 
\end{equation}
${W}_k$ denotes the concept prototype of class $k$. We use cosine similarity to measure the relationships and construct a similarity confusion matrix. Using the unstructured FGDG model (without concept or feature structuralization) as a baseline, we compare the prototype similarity patterns of both FGDG and CFSG under the multi-granularity setting (Fig. \ref{allfgdg} and \ref{allCFSG}). Compared to the ground-truth similarity matrix derived from hierarchical labels (Fig. \ref{truth}), the prototype similarity produced by CFSG (Fig. \ref{allCFSG}) shows strong consistency. Furthermore, we compute the Spearman rank correlation between the model-derived similarities and the ground-truth similarities across the entire concept space, as well as separately for the common, specific, and confounding regions. As shown in Table \ref{tab:Ground Truth}, the CFSG model achieves a high correlation of 0.97, substantially outperforming the unstructured FGDG baseline, which only achieves 0.69. The common, specific, and confounding components in the CFSG model achieve Spearman correlations of 0.97, 0.97, and 0.91, respectively, showing substantial improvements over the FGDG model. These results demonstrate that the structured concept space learned by CFSG effectively captures inter-class similarity and integrates multi-granularity structural knowledge into both the concept and feature spaces. Moreover, they validate that feature structuralization induces structural patterns within the concept space.

\begin{table}[!t]
\centering
\begin{tabular}{cccc}
\hline
\multicolumn{4}{c}{Category Examples} \\
\hline
g=0 & g=1 & g=2 & g=3 \\
\hline
8     & 5     & 3     & 3 \\
9     & 6     & 3     & 3 \\
10    & 5     & 3     & 3 \\
11    & 7     & 3     & 3 \\
12    & 8     & 3     & 3 \\
28    & 19    & 12    & 3 \\
29    & 19    & 12    & 3 \\
51    & 36    & 19    & 8 \\
\hline
\end{tabular}
\caption{Labels of fine-grained categories across different granularity levels.}
\label{tab:label}
\end{table}

\begin{table}[!t]
\centering
\begin{tabular}{c   |cccc}
\hline
FGDG & All & Com. & Spe. & Conf.\\
\hline
Ground Truth & 0.69 & 0.70 & 0.67 & 0.67 \\
\hline
CFSG & All & Com. & Spe. & Conf.\\
\hline
Ground Truth & 0.97 & 0.97 & 0.97 & 0.91 \\
\hline
\end{tabular}
\caption{Spearman's rank correlation between region-level concepts and ground-truth labels for each model.}
\label{tab:Ground Truth}
\end{table}

\end{document}